\documentclass[letterpaper, 10 pt, conference]{ieeeconf}  

\IEEEoverridecommandlockouts                              

\usepackage{graphics} 
\usepackage{epsfig} 
\usepackage{mathptmx} 
\usepackage{times} 
\usepackage{amsmath} 
\usepackage{amssymb}  
\usepackage{todonotes}
\usepackage{etoolbox}

\usepackage[font=small,labelfont=bf]{caption} 

\usepackage{float}
\usepackage{hyperref}
\usepackage{sidecap}
\usepackage{lipsum} 

\title{\LARGE \bf
ImaginationPolicy: Towards Generalizable, Precise and Reliable End-to-End Policy for Robotic Manipulation
}

\author{Dekun Lu$^{*}$, Wei Gao$^{*}$ and Kui Jia%
\thanks{$^{*}$Equal contribution; more junior authors listed earlier.}%
}

\begin{document}
\maketitle
\thispagestyle{empty}
\pagestyle{plain}

\begin{abstract}

End-to-end robot manipulation policies offer significant potential for enabling embodied agents to understand and interact with the world. Unlike traditional modular pipelines, end-to-end learning mitigates key limitations such as information loss between modules and feature misalignment caused by isolated optimization targets. Despite these advantages, existing end-to-end neural networks for robotic manipulation—including those based on large vision-language-action (VLA) models—remain insufficiently performant for large-scale practical deployment. In this paper, we take a step towards an end-to-end manipulation policy that is generalizable, accurate and reliable. 
To achieve this goal, we propose a novel Chain of Moving Oriented Keypoints (CoMOK) formulation for robotic manipulation. Our formulation is used as the action representation of a neural policy, which can be trained in an end-to-end fashion (details in Sec.~\ref{sec:method}). Such an action representation is general, as it extends the standard end-effector pose action representation and supports a diverse set of manipulation tasks in a unified manner.
The oriented keypoint in our method enables natural generalization to objects with different shapes and sizes, while achieving sub-centimeter accuracy.
Moreover, our formulation can easily handle multi-stage tasks, multi-modal robot behaviors, and deformable objects.
Extensive simulated and hardware experiments demonstrate the effectiveness of our method.
Video demonstration, source code and supplemental materials are available on \href{https://sites.google.com/view/imaginationpolicy}{\textcolor{blue}{\underline{https://sites.google.com/view/imaginationpolicy}}}

\end{abstract}

\section{Introduction}
\label{sec:intro}

During the past decade, deep learning has profoundly reshaped the landscape of robotic manipulation. One popular approach leverages deep neural networks (DNNs) as functional submodules in traditional modularized pipelines. This significantly enhances the capabilities of robot manipulators, particularly in robot perception~\cite{zeng2017multi, tremblay2018deep, sahin2018category}. Concurrently, a distinct paradigm has emerged, pursuing end-to-end, pixel-to-torque learning strategies~\cite{chi2023diffusion, levine2016end, zhu2018reinforcement}. This alternative methodology aims to circumvent the information bottleneck between discrete modules in traditional pipelines by directly mapping raw sensory input to low-level control actions (e.g., robot joint torque commands).

The success of Large Language Models (LLMs)~\cite{achiam2023gpt, bai2023qwen, touvron2023llama} and Vision-Language Models (VLMs)~\cite{wang2024qwen2vl, liu2023visual, liu2024improved} over the past years has injected new inspiration into robotic manipulation research. A key insight driving this interest is the promising generalization capability of LLMs/VLMs, benefits from large-scale, multiple-task pretraining. Motivated by this potential, researchers are increasingly exploring Vision-Language-Action (VLAs) models~\cite{kim2024openvla, black2024pi0, liu2024rdt} – aiming to imbue robots with similar generalization capability. For robotic manipulation, this desired generalization manifests primarily in two critical dimensions: 1) generalization across scenes and objects (robustly handling novel objects with variations on shape, size, and appearance), and 2) generalization across diverse manipulation tasks (adapting to new instructions and tasks without expensive retraining).

Despite significant research advances, current VLA models have not yet seen large-scale deployment in real-world industrial or service applications. Their performance in key operational perspectives, such as reliability and accuracy, can be insufficient for practical deployment and sometimes inferior to well-engineered traditional modularized pipelines. This shortcoming becomes particularly pronounced when these VLAs are deployed in unseen objects or different robot platforms. Consequently, bridging the gap between the potential for broad generalization and the practical demands of real-world performance remains a central challenge. 

In this paper, we propose an alternative approach for end-to-end robot manipulation that emphasizes reliability, accuracy, and interpretability.
To achieve this, we propose a novel affordance-based formulation for robotic manipulation, where we define affordance as task-specific, semantically meaningful local object parts.
To translate this cognitive concept into executable robotic actions, we propose to represent affordance as semantic, task-specific oriented keypoint(s). Subsequently, the manipulation behavior can be represented as a Chain of Moving Oriented Keypoints (CoMOK), as detailed in Sec.~\ref{sec:method}.
This affordance-base formulation is used as the action representation of a neural network based manipulation policy, which can be trained end-to-end.
This action representation is general, as it reduces to the standard end-effector pose action in a special case. The proposed formulation provides a unified framework for diverse manipulation tasks, including ones previously studied in isolation (such as robot grasping algorithms).
The oriented keypoint affordance enables natural generalization to objects with different shape and size, while achieving sub-centimeter level accuracy. Moreover, the proposed formulation can easily handle multi-stage tasks, multi-modality robot behaviors and deformable objects.

We also design a new neural network architecture based on the proposed formulation. The network takes sensory observation and task encoding (e.g., language prompt) as inputs. It uses score-matching~\cite{urain2022se3, chi2023diffusion} (a variant of diffusion model) to produce the multi-modal robot actions. The network can be trained end-to-end and the output can be directly used for joint-level motion generation. This joint-level motion generator can be either 1) a traditional motion planning algorithms; or 2) a neural-network based trajectory generation pipeline (which can be trained end-to-end). We attempt both types of motion generators on simulated and real-world experiments. Our network demonstrates promising generalization capability without sacrificing accuracy/reliability.

This paper is organized as follows: in Sec.~\ref{sec:related} we review related works. Sec.~\ref{sec:method} describes our manipulation formulation and several important extensions. Sec.~\ref{sec:impl} introduces the diffusion-based neural network to address multi-modal, multi-stage manipulation tasks. Sec.~\ref{sec:results} demonstrates our methods on several simulated and real world tasks and shows generalization of our method. Sec.~\ref{sec:conclusion} concludes. 

\section{Related Works}
\label{sec:related}

\subsection{Robot grasp detection}
\label{subsec:grasping}

Grasping algorithms enable finding stable grasp poses that allow robots to reliably pick up objects. Grasping detection is extensively studied~\cite{zeng2018affordance, gualtieri2016gpd, morrison2018closing, mahler2019learning}, please refer to~\cite{du2021vision} for a comprehensive review. 
%
%
In this work, we focus on a general action representation for robotic manipulation. Grasp detection would be one task within the proposed formulation.

\subsection{Affordance-based robot manipulation}

Affordance is traditionally used as a cognitive concept to investigate the behaviors of humans and animals~\cite{osiurak2017affordance}. In the context of robotic manipulation, affordance typically refers to local parts (of an object) that are relevant to some robot manipulation tasks. Thus, some researchers~\cite{myers2015affordance, kokic2017affordance, li2024laso, tang2025uad} try to identify these ``affordance'' regions from image or point clouds using object detection and segmentation neural networks. In addition to the cognitive understanding, the concept of affordance can be used to construct executable robot actions. In particular, a series of contributions~\cite{gao2019kpam, qin2019keto, gao2021kpam2} are built upon keypoint-based object and scene representation. This keypoint representation has been integrated with various techniques, such as imitation learning~\cite{wang2025skil}, constraint optimization~\cite{huang2024rekep, wong2021manipulation} and LLM-based reasoning~\cite{huang2024rekep, pan2025omnimanip}. With keypoint representation, these methods have demonstrated promising generalization with respect to different environments and manipulated objects.

Our method is also based on keypoint affordance mentioned above. Compared with these works, the key distinction is the generalization of the proposed formulation. Existing works are typically limited to particular tasks (e.g., object placement and tool usage), while the proposed method aims at a general action representation. Various manipulation skills (e.g., grasping detection reviewed in Subsec.~\ref{subsec:grasping}) can be regraded as a special case of our formulation, which is out of the scope of existing keypoint-based manipulation pipelines. Moreover, existing methods are typically restricted to rigid objects, while our method can handle deformable objects.


\subsection{End-to-end learning for robotic manipulation}

In an end-to-end policy, the entire manipulation pipeline can benefit from joint optimization. This might lead to performance improvement compared to traditional modularized manipulation pipelines, where each module (e.g., perception, motion planning and control) is implemented/trained in isolation. Most contributions on end-to-end robotic manipulation~\cite{finn2016deep, chi2023diffusion, levine2016end, zhu2018reinforcement} would train neural networks that directly map raw sensory input to low-level control actions. These networks are typically trained with imitation learning~\cite{chi2023diffusion, zhao2023act, ke20243d} and/or reinforcement learning~\cite{schulman2015trust, schulman2017proximal, levine2016contact, van2016stable}. Besides, many works build VLA~\cite{kim2024openvla, black2024pi0, liu2024rdt} by incorporating end-to-end manipulation policy learning with VLMs, which excels at visual/language understanding due to large-scale pre-training.

The proposed method falls into this category. Our method differs from these existing works mainly in terms of the action representation. Existing work typically produces end-effector poses or robot joint angles. In this paper, we propose a novel affordance-based action representation to enable better generalization and accuracy. Moreover, our action representation is general, as it falls back to the traditional end-effector pose action as a special case.

\section{Affordance-based Action Representation}
\label{sec:method}

In this section, we present the proposed formulation for robotic manipulation. We would first discuss the method in the simplest setup, then make several extensions.

\begin{figure}[t]
\centering
\includegraphics[width=0.4\textwidth]{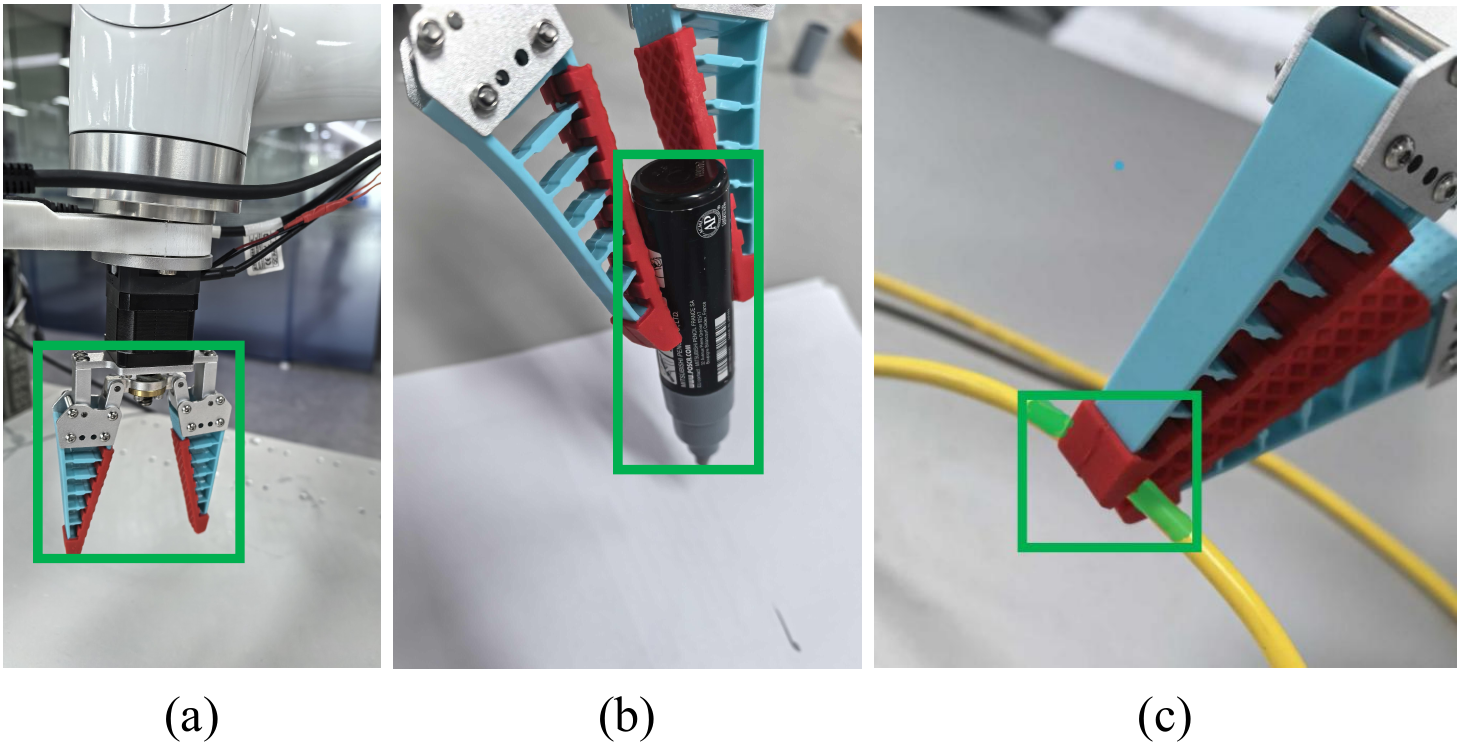}
\caption{\label{fig:control} An illustration of the robot's control authority in the proposed formulation. (a) The robot can command 6-DoF movements over its end-effector. (b) The robot can command 6-DoF movements of a rigid object grasped by the robot. (c) The robot can command 6-DoF movements of a \textit{local patch} (illustrated in green mask) of a deformable object grasped by the robot.
}
\end{figure}

Consider an environment in which some manipulation tasks should be performed. The robot is capable of performing manipulation actions with respect to part of the environment, in particular:

\begin{itemize}
    \item The robot has full control authority over its end-effector (which is regarded as part of the environment), as shown in Fig.~\ref{fig:control} (a).
    \item If a rigid object is held by the robot, then the robot has full control authority of that object. In other words, the robot can command the 6-DoF movement of that object (subject to physical constraints such as reachability or collision). An illustration is shown in Fig.~\ref{fig:control} (b). 
    \item If a deformable object is grasped by the robot, then the robot does not have full control authority of that object (since the object may have more than 6 DoFs). On the other hand, the robot can command the 6-DoF movement of a \textit{local grasped patch} of that object, as illustrated in Fig.~\ref{fig:control} (c).
\end{itemize}

An illustration is provided in Fig.~~\ref{fig:control}. For an object that is not yet picked-up, the robot obviously has no control over it. However, the robot can ``imagine'' its control authority of that object and the manipulation behavior with that object in hand, if the robot plans to pick up that object.

With this representation, we propose to train the following (abstract form of) neural network. The neural network takes scene observation $o_\text{scene}$ and task description $f_\text{task}$ as input. The scene observation $o_\text{scene}$ might be point clouds ($R^{M \times 3}$) or images ($R^{H \times W \times 3}$). The task description $f_\text{task}$ might be natural language (e.g., ``pick up the mug''). In the simplest setup, the output of the neural network is formulated as
\begin{equation}
\label{equ:network}
    \text{network}(o_\text{scene}, f_\text{task}) \xrightarrow{} (o_\text{manipulated}, T_\text{affordance}, T_\text{action})
\end{equation}

\noindent where $o_\text{manipulated}$ is the part of the environment over which the robot has or would have some control authority, for the task $f_\text{task}$. Practically, $o_\text{manipulated}$ might be represented as a bounding box or segmentation mask of environment $o_\text{scene}$.

In the simplest setting, the output $T_\text{affordance}, T_\text{action} \in SE(3)$ are 6 DoF frames, their semantics are defined as

\begin{itemize}
    \item $T_\text{affordance}$ represents a task-related keypoint affordance defined with respect to $o_\text{manipulated}$. It contains the orientation information (similar to oriented keypoints in kPAM 2.0~\cite{gao2021kpam2}). For rigid objects, $T_\text{affordance}$ is rigidly attached to the $o_\text{manipulated}$ object; for deformable objects, $T_\text{affordance}$ is rigidly attached to a local patch of the picked (or to be picked) object.
    \item $T_\text{action}$ implies: if the $T_\text{affordance}$ mentioned above is manipulated to be aligned with $T_\text{action}$, then the task described by $f_\text{task}$ is accomplished.
\end{itemize}

In the following subsections, we explain the semantics of the network with a detailed example in Subsec.~\ref{subsec:example}. Then, we extends the simplest setup in Equ.~\ref{equ:network} into multi-stage manipulation (Sec.~\ref{subsec:multistage}), multi-modal manipulation behaviors (Sec.~\ref{subsec:multimodal}) and trajectory actions (Sec.~\ref{subsec:trajaction}).

\subsection{An Illustrative Example}
\label{subsec:example}

\begin{figure}[t]
\centering
\includegraphics[width=0.45\textwidth]{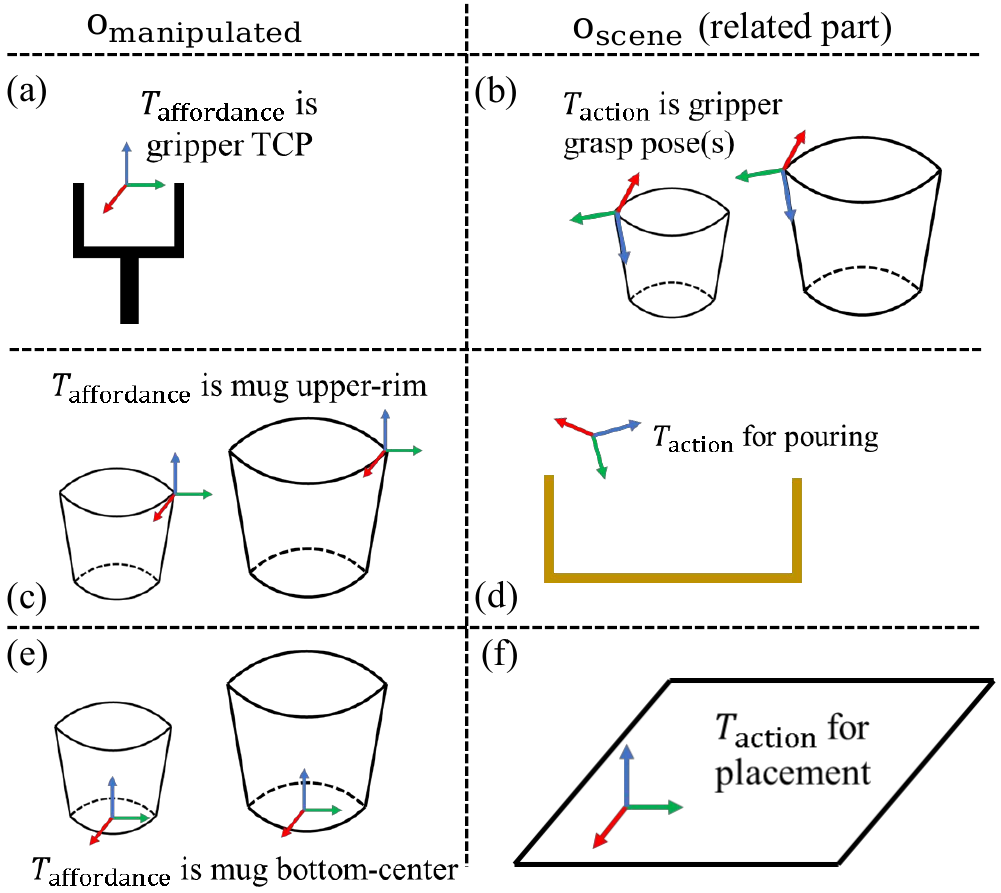}
\caption{\label{fig:pouring} An illustration of the proposed formulation using the water-pouring task as an example, which involves three sub-tasks: 1) the robot picks up the mug using its gripper, as shown in (a, b); 2) the robot manipulates the picked mug to pour the water, as shown in (c, d); and 3) the robot places the mug onto the table, as shown in (e, f). Each stage has its distinct, task-related definition of objects $o_\text{manipulated}$ and affordance-action pairs $T_\text{affordance},~T_\text{action}$. We draw two mug instances to highlight the generalization to different objects.
}
\end{figure}

We use the following water-pouring application to demonstrate the formulation, as shown in Fig.~\ref{fig:pouring}. This task consists of three stages, and in the simplest setting it requires three task description for each stage. They are $f_\text{task\_1}$: the robot picks up the mug; $f_\text{task\_2}$: the robot manipulates the mug to pour the water into a container; and $f_\text{task\_3}$: the robot places the mug onto a table. Moreover, we would like the manipulation skill to be generalizable with respect to different mugs, as illustrated in Fig.~\ref{fig:pouring}.

These three (sub)tasks involve different network output ($o_\text{manipulated}$, $T_\text{affordance}$ $T_\text{action}$), as detailed below:

\begin{itemize}
    \item In stage 1), the robot picks up the mug, as shown in Fig.~\ref{fig:pouring} (a) (b). The $o_\text{manipulated}$ is the robot gripper tool, and $T_\text{affordance}$ is the tool TCP frame. The $T_\text{action}$ is the grasp pose for the mug in $o_\text{scene}$. This is a standard grasp pose detection problem.
    \item In stage 2), the robot moves the mug to pour the water, as shown in Fig.~\ref{fig:pouring} (c) (d). The $o_\text{manipulated}$ becomes the mug at this stage, and the $T_\text{affordance}$ is at the upper-rim of the mug. The $T_\text{action}$ is pouring configuration of the mug with respect to a container in the environment $o_\text{scene}$ where the water should be poured.
    \item In stage 3), the robot places the mug onto the table after pouring, as shown in Fig.~\ref{fig:pouring} (e) (f). The $o_\text{manipulated}$ is again the mug, and the $T_\text{affordance}$ becomes the mug bottom-center at this stage. The $T_\text{action}$ is a stable placement configuration for the mug bottom-center, on a table in the environment $o_\text{scene}$.
\end{itemize}

As illustrated in Fig.~\ref{fig:pouring}, this formulation works for mugs with different shapes and sizes, supposed the network predicts correct $T_\text{affordance}$ and $T_\text{action}$. This generalization to new objects comes from these oriented keypoints capture task-specific local geometric information of related objects, while ignoring irrelevant geometric details (similar to the idea of kPAM~\cite{gao2019kpam, gao2021kpam2}).
Moreover, if $o_\text{manipulated}$ is always fixed to the gripper tool (and $T_\text{affordance}$ is the tool TCP frame), the proposed formulation falls back to the usual end-effector pose action representation.

\subsection{Extension to Multi-Stage Manipulation}
\label{subsec:multistage}

As illustrated in the water-pouring example in Subsec.~\ref{subsec:example}, a long-horizon manipulation task might involve several sub-tasks, which requires the switching of $f_\text{task}$ in formulation Equ.~\ref{equ:network}. Although manually setup the corresponded $f_\text{task}$ for each sub-task is feasible, an alternative approach is to perform automatic decision of sub-task from a global task description $f_\text{task\_global}$. Using Fig.~\ref{fig:pouring} as an example, given a global task description $f_\text{task\_global}$ ``pour water into the container'', the network automatic decide three sub-tasks $f_\text{task\_1}$, $f_\text{task\_2}$ and $f_\text{task\_3}$ mentioned above. This decision-making also takes the environment observation $o_\text{scene}$ as input, for instance $f_\text{task\_1}$ ``grasp the mug'' is omitted if the mug is already picked up. Formally, the formulation for multi-stage manipulation is
\begin{equation} \label{equ:stage_network}
\begin{split}
 &\text{network}(o_\text{scene}, f_\text{task\_global}) \xrightarrow{} \text{List}[a_\text{stage}] \\
 \text{where~}& a_\text{stage} = (f_\text{task\_stage}, o_\text{manipulated}, T_\text{affordance}, T_\text{action}) \\
\end{split}
\end{equation}

Automatic generation of sub-task descriptions $f_\text{task\_stage}$ from a global task description $f_\text{task\_global}$ has been explored in several existing works. These methods typically prompt an existing LLM/VLM~\cite{liang2022code,wang2023robogen} or perform finetuning of LLM/VLM using customized task-planning datasets~\cite{zawalski2024robotic}. The proposed formulation augment these existing works with new action representation $T_\text{affordance}$ and $ T_\text{action}$.

\subsection{Extension to Multiple Action Candidates}
\label{subsec:multimodal}

The formulation in Sec.~\ref{sec:method} only produces one action instance $(o_\text{manipulated}, T_\text{affordance}, T_\text{action})$ for the input. Practically, there might be many action candidates that can accomplish the task for the input. This multi-modality can come from different perspectives, two important cases are: 1) different $o_\text{manipulated}$ instances; and 2) different $T_\text{affordance}$ and $ T_\text{action}$ pairs for a given $o_\text{manipulated}$ instance. Using the pour water in Fig.~\ref{fig:pouring} as an example, there might be more than one mugs and/or containers in the environment. Moreover, the grasp detection in Fig.~\ref{fig:pouring} (a, b) might permit many feasible grasp poses for a mug instance; and there might be many stable placement poses in Fig.~\ref{fig:pouring} (e, f) on the table.


These different perspectives of multi-modality can be characterized with neural networks. For multiple instances of $o_\text{manipulated}$ in the scene, standard object detection and segmentation network can address it. For different $T_\text{affordance}$ and $ T_\text{action}$ pairs, one approach is to use diffusion models~\cite{chi2023diffusion, urain2022se3} to characterize this action distributions. Please refer to Sec.~\ref{sec:impl} for a detailed description of our implementation.

\begin{figure}[t]
\centering
\includegraphics[width=0.35\textwidth]{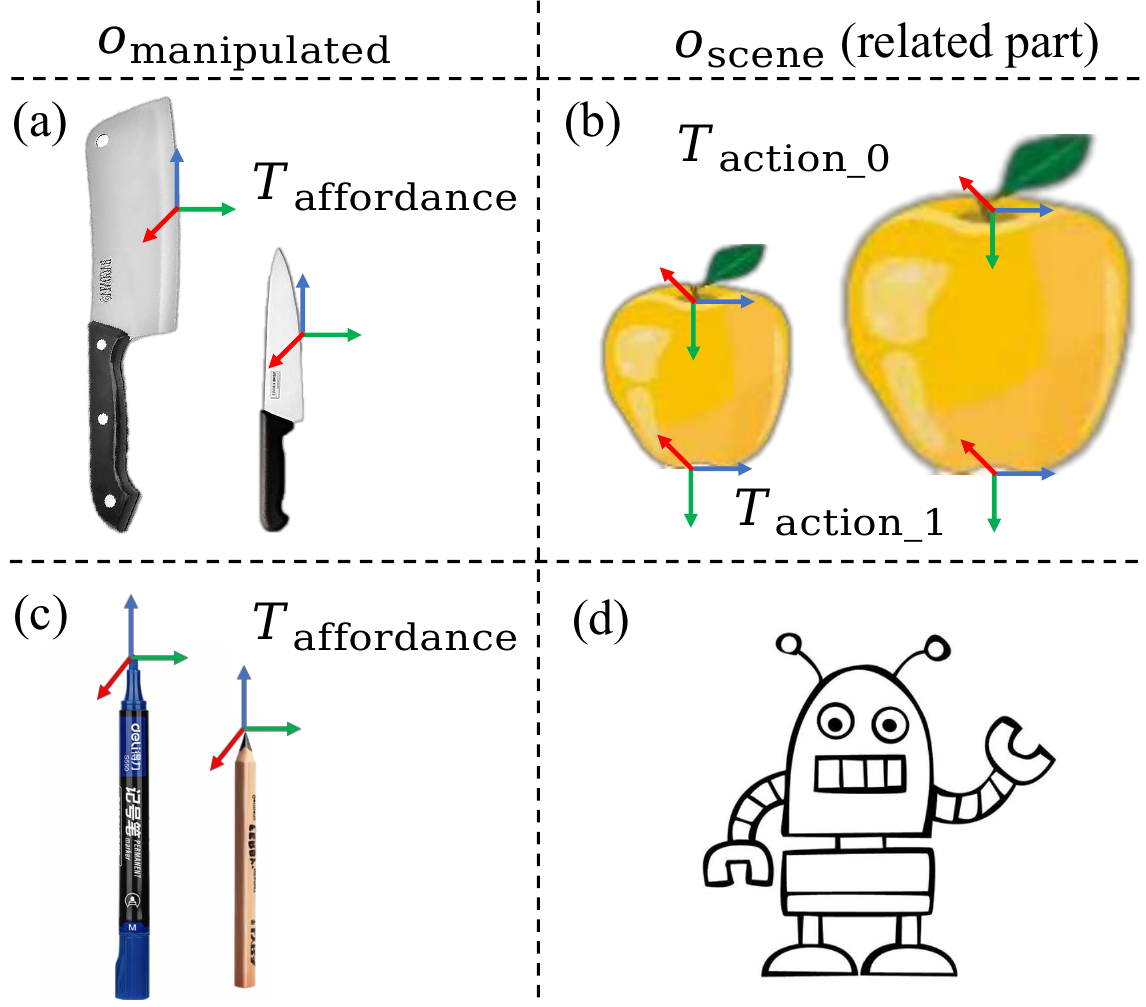}
\caption{\label{fig:cutting} An illustration of the trajectory action in Subsec.~\ref{subsec:trajaction}. (a) (b) The trajectory might be sparse, for instance the fruit-cutting task might involve two action frames: the $T_\text{action\_0}$ above the fruit and $T_\text{action\_1}$ below it. (c) (d) The trajectory might be dense, for instance the drawing task might involve hundreds of $T_\text{action}$ frames.
}
\end{figure}

\subsection{Extension to Trajectory Action}
\label{subsec:trajaction}

In the original formulation Equ.~\ref{equ:network}, the action $T_\text{action}$ is one SE(3) set-point of the $T_\text{affordance}$ frame. Practically, various manipulation tasks cannot be accomplished with only one set-point. As shown in Fig.~\ref{fig:cutting} (a, b), the manipulation task ``cutting fruit with a knife'' requires a simple top-down movement of the knife. This movement can be described by a $T_\text{affordance}$ frame attached to the blade with two $T_\text{action\_0}$ and $T_\text{action\_1}$, which implies the $T_\text{affordance}$ frame must be moved aligned with $T_\text{action\_0}$ at first, then move to $T_\text{action\_1}$ to cut the fruit. For a more complex manipulation tasks, such as robot welding/drawing, the trajectory should be described by a dense sequence of $T_\text{action}$, as shown in Fig.~\ref{fig:cutting} (c, d).

Formally, the original formulation Equ.~\ref{equ:network} can be extended to a sequence of SE(3) frames trajectory 
\begin{equation} \label{equ:stage_network}
\begin{split}
 \text{network}(o_\text{scene}, f_\text{task}) &\xrightarrow{} (o_\text{manipulated}, T_\text{affordance}, T_\text{action\_seq}) \\
 T_\text{action\_seq}& \in SE(3) \times K \\
\end{split}
\end{equation}
\noindent where $K$ is the time horizon. In this formulation, the semantics of $T_\text{affordance}$ is the same as Equ.~\ref{equ:network}. The semantics of $T_\text{action\_seq}$ is: if the $T_\text{affordance}$ is manipulated to move along the $SE(3)$ trajectory $T_\text{action\_seq}$, then the task defined by $f_\text{task}$ is accomplished.

An interesting design decision is whether the initial point $T_\text{action\_0}$ of the $T_\text{action\_seq}$ trajectory must be aligned with the current $T_\text{affordance}$. If it is required, then the $T_\text{affordance}$ and $T_\text{action\_seq}$ can be converted into a trajectory of end-effector, which can be directly executed on the robot. Moreover, the network can serve as a reactive manipulation policy if it is evaluated online with some frequency (e.g., 10 Hz).

If the initial alignment not required, for instance in Fig.~\ref{fig:cutting} (a, b) the ``pre-cut'' point $T_\text{action\_0}$ might not be aligned with initial $T_\text{affordance}$, then the movement from $T_\text{affordance}$ to $T_\text{action\_0}$ can be thought as ``irrelevant'' to the manipulation task $f_\text{task}$. In other words, any physically feasible robot joint-space trajectories (subject to collision, reachability and other constraints) that moves the initial $T_\text{affordance}$ to $T_\text{action\_0}$ would satisfy the requirement of $f_\text{task}$. Using Fig.~\ref{fig:cutting} as an example, the task $f_\text{task}$ ``cut the apple'' only requires the knife blade to perform a cutting movement from $T_\text{action\_0}$ to $T_\text{action\_1}$, any robot joint-space trajectory that moves the blade to $T_\text{action\_0}$ would be feasible for the task.

To generate the trajectory from $T_\text{affordance}$ to $T_\text{action\_0}$, we first compute end-effector pose target from $T_\text{affordance}$ to $T_\text{action\_0}$. Traditional motion planning algorithms~\cite{noreen2016optimal} or learning-based approaches~\cite{fishman2023motion} can be used to generate the robot joint-space trajectory. Moreover, if more than one $T_\text{action\_seq}$ candidates are produced by the neural network, as mentioned in Subsec.~\ref{subsec:multimodal}, then the motion generator can select one candidate that is both physical feasible and optimal (e.g., in terms of joint-space distance). We have explored both traditional and learning-based methods in our implementation, as detailed in Sec.~\ref{sec:impl}.

\begin{figure}[t]
\centering
\includegraphics[width=0.47\textwidth]{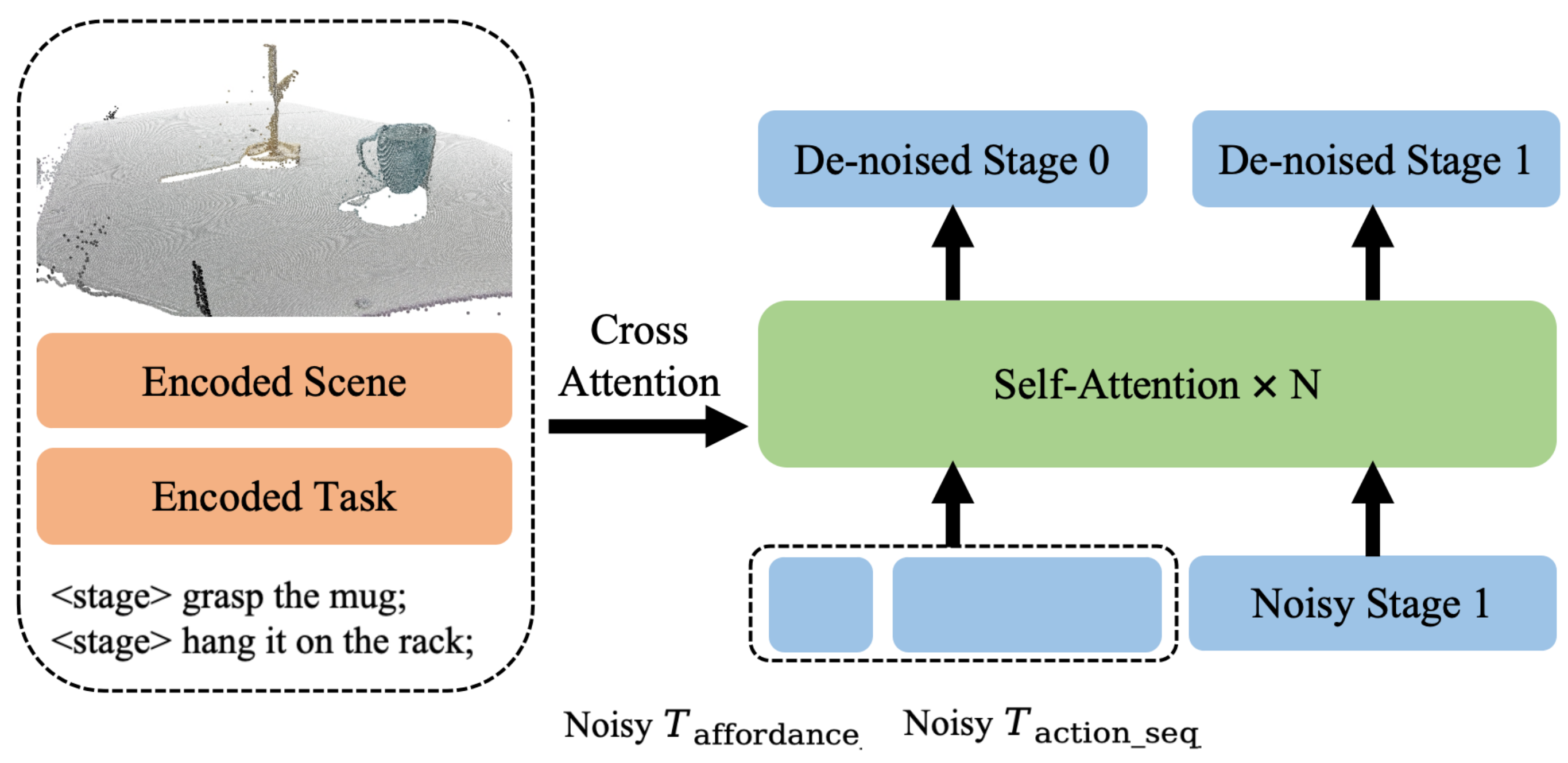}
\caption{\label{fig:network} An illustration of the action prediction network in Sec.~\ref{sec:impl}. The network takes scene observation (point cloud) and task description as input. We use a score-matching network to characterize the multi-modal distribution of robot action candidates, as described in Subsec.~\ref{subsec:multimodal}. Please refer to Sec.~\ref{sec:impl} for a detailed description.
}
\end{figure}

\section{Neural Network Architecture and Implementation Detail}
\label{sec:impl}

\vspace{0.5em}
\noindent \textbf{Network Architecture.} Based on the formulation above, we design a multi-stage neural network architecture that maps raw sensor observation into robot actions, as detailed below.

The task-planning network takes RGBD images and task description $f_\text{task\_global}$ as input, and it produces a list of sub-tasks and the corresponded $o_\text{manipulated}$. Optionally, the network also produces an $o_\text{env}$ detection which indicates the region that $T_\text{action}$ should ``pay attention to''. Both $o_\text{manipulated}$ and $o_\text{env}$ is used as attention masks to produce $T_\text{affordance}$ and $T_\text{action}$ subsequently. Formally,

\begin{equation} \label{equ:task_network}
\begin{split}
 \text{task\_planning} &(o_\text{scene\_RGB}, f_\text{task\_global}) \xrightarrow{} \text{List}[t_\text{stage}] \\
 \text{where~}& t_\text{stage} = (f_\text{task\_stage}, o_\text{manipulated}, o_\text{env}) \\
\end{split}
\end{equation}

\noindent We fine-tune a Groma~\cite{ma2024groma} VLM for the task-planning network, which takes image and language prompt as input and it produces both language output and region detection.

The action-prediction network takes scene point cloud and a list of staged task $(h_\text{task\_stage}, o_\text{manipulated}, o_\text{env})$ as input, where $h_\text{task\_stage}$ is the network-encoded hidden features of $f_\text{task\_stage}$ from previous stage. The scene point cloud is computed from the depth channel of the input RGBD images, except for the grasp detection task in Sec.~\ref{sec:results}. For the grasp detection experiment, we use the point cloud dataset from an existing work~\cite{urain2022se3}, which only contains the object of interest. Obviously the grasp detection has only one stage, with both global and staged task descriptions be ``pick up the \{object$\_$label\}''.

The action-prediction network predicts the $T_\text{affordance}$ and $T_\text{action\_seq}$ for all stages jointly. To handle multiple action candidates as described in Subsec.~\ref{subsec:multimodal}, we use a score-matching network (a variant of diffusion model) inspired by~\cite{urain2022se3} that performs iterative denoising from initial noisy input. A transformer-based architecture is used in our implementation, as shown in Fig.~\ref{fig:network}. The point cloud is first encoded to obtain a sequence of features. Tokens corresponded to noisy $T_\text{affordance}$ and $T_\text{action\_seq}$ are flattened and concentrated. A series of self-attention layer is used to predict the denoising vector for each $SE(3)$ elements in $R^6$, similar to~\cite{urain2022se3}. Moreover, a series of cross attention layer is used to incorporate features from point cloud and task description, with optional attention masks of each stage.
For tasks with less stages or action sequence length, we use attention masks to avoid attention and loss-computing on them. For the action-prediction network, we do not back-propagate the gradient through $h_\text{task\_stage}$. As an important special case, for some manipulation tasks the $o_\text{manipulated}$ becomes the robot gripper, while the $T_\text{affordance}$ is the TCP frame of the gripper. We do not explicit predict it using the neural network, instead set the $T_\text{affordance}$ target to identity.

\vspace{0.3em}
\noindent \textbf{Robot Trajectory Generation.} We explore two motion generation methods given the output of action prediction mentioned above. The output $T_\text{affordance}$ and $T_\text{action\_seq}$ are first converted into robot end-effector space trajectory candidates for each stage. For simulated experiment, we use a learned policy~\cite{fishman2023motion} to generate robot joint-space trajectories. This policy network takes encoded scene point cloud and picked object point cloud as inputs, and it produce joint-space displacement that would move the robot to the target.

For real-world experiment, we use a traditional task-and-motion-planning algorithm implemented in~\cite{gao2024kilobot}, due to the unsatisfactory performance of~\cite{fishman2023motion} in real-world. The task-and-motion-planning algorithm~\cite{gao2024kilobot} automatically perform selection from multiple candidates based on physical constraints such as collision, reachability and trajectory length.

\section{Results}
\label{sec:results}

In this section, we demonstrate a variety of manipulation tasks using the proposed formulation. The particular novelty of these demonstrations is that our network can simultaneously: 1) address different the manipulation tasks in a unified framework; and 2) handle large shape variations of objects in each tasks. We utilize a 6-DOF robot arm (Rokae SR5) mounted with a parallel jaw gripper. An RGBD sensor is also mounted on the end effector. The video demo on our \href{https://sites.google.com/view/imaginationpolicy}{\textcolor{blue}{\underline{project page}}} best demonstrates our solution to these tasks. 

\begin{figure*}[t]
\centering
\includegraphics[width=0.87\textwidth]{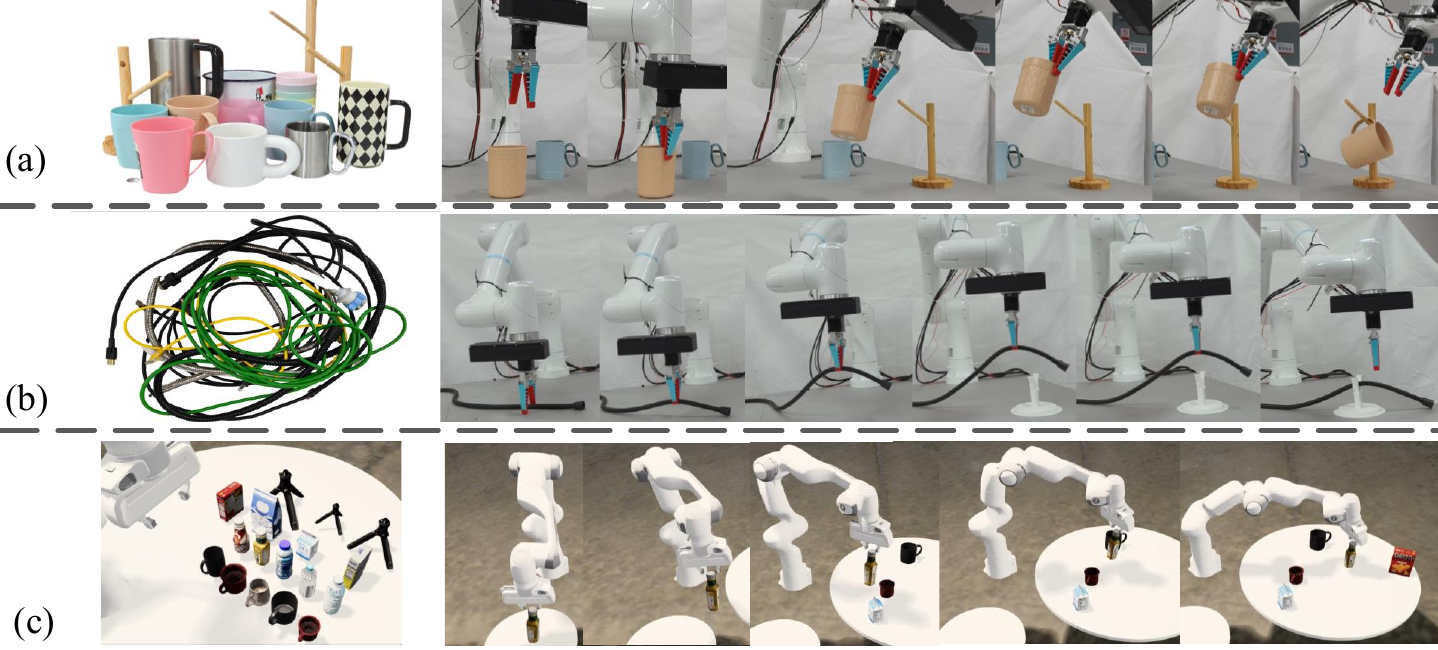}
\caption{\label{fig:experiment} An overview of our experiments. The tasks are: (a) hang mugs on a rack by the mug handles; (b) insert cables into a clamps; (e) place object stably onto a table. Experiment (a) (b) are conducted in real-worlds while experiment (c) is carried out in simulation. The particular novelty of these demonstrations is that our network can simultaneously: 1) address different the manipulation tasks in a unified framework; and 2) handle large shape variations of objects in each tasks. }
\end{figure*}

\begin{figure}[t]
\centering
\includegraphics[width=0.45\textwidth]{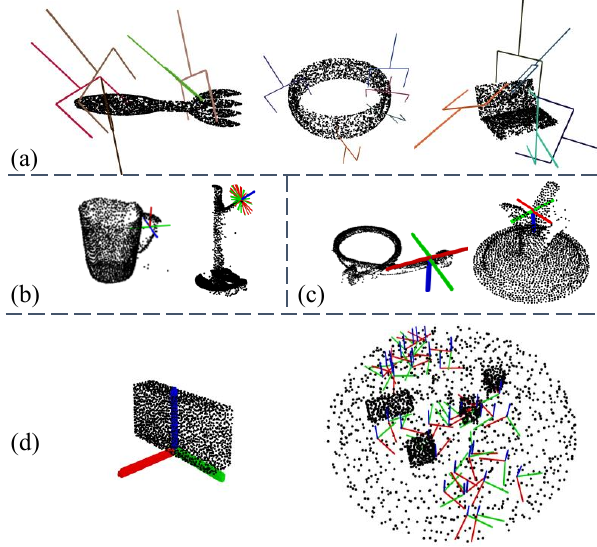}
\caption{\label{fig:action_output} A visualization of the predicted action distribution. For the grasping task (a), only the $T_\text{action}$ (which is the detected grasp pose) is visualized. For (b-d), both $T_\text{affordance}$ and one representative frame of $T_\text{action\_seq}$ are visualized. Please refer to Sec.~\ref{sec:results} for more details.
}
\end{figure}

\begin{figure}[t]
\centering
\includegraphics[width=0.4\textwidth]{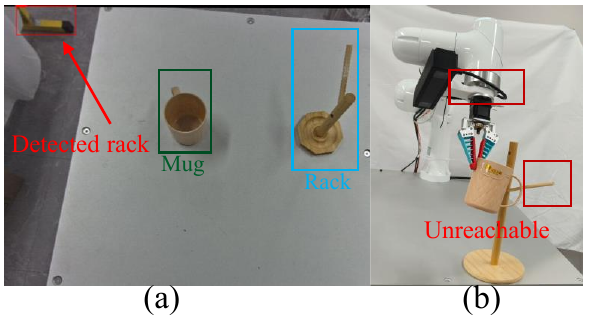}
\caption{\label{fig:failure} Several failure modes of task execution. (a) Detection failure of the task-planning network. (b) Reachability limitation that is hard to be taken into consideration during motion generation (restricted by the camera wire in this example).
}
\end{figure}

\subsection{Task Description}
\label{subsec:task_descri}

\noindent \textbf{Parallel-jaw grasping: } We first conduct a proof-of-concept experiment of grasp poses detection for a parallel-jaw gripper. We use the dataset from an existing work~\cite{urain2022se3}. In addition to the actual grasp pose, the network would produce another ``pre-grasp'' pose, which is at some distance from the grasp pose along the $z$-axis of the gripper.

This task has only one stage. The $o_\text{manipulated}$ for this task is the robot gripper tool (which does not appear in the $o_\text{scene}$), while $T_\text{affordance}$ should be the gripper TCP frame. As mentioned in Sec.~\ref{sec:impl}, the $T_\text{affordance}$ frame is not explicitly predicted by the network if $o_\text{manipulated}$ is the gripper. The meaningful output from the network is only $T_\text{action\_seq}$, which consists of two frames (``pre-grasp'' and ``grasp''). 

\vspace{0.3em}
\noindent \textbf{Stable placement: } We perform a simulated experiment as shown in Fig.~\ref{fig:experiment} (c). the robot is expected to pick-up an object (from a small table) and place it onto a (large) table.  We use three categories of objects in this task: bottle, box and tripod. For each object category, we use multiple instances of objects with vastly different shapes, sizes and appearances, as shown in Fig.~\ref{fig:experiment} (c). In particular, we use 20 bottles, 10 boxes and 5 tripods in our experiment. Moreover, we randomly drop about 5 objects on the table as obstacles. Successful robot placement actions should satisfy: 1) it is stable; 2) the placed object does not penetrate into the table; and 3) the placed object does not collide with obstacle objects.

This task has two stages. In stage 1, the robot would pick up the object, and $o_\text{manipulated}$ for this stage is the gripper tool. In stage 2, the robot move the object to a suitable placement configuration and place it on the table. The $o_\text{manipulated}$ for this stage is the picked object. The $T_\text{affordance}$ frame of stage 2 implies a configuration of the picked object that can be stably placed on a flat surface, as shown in Fig.~\ref{fig:action_output} (d). The $T_\text{action\_seq}$ in stage 2, which consists of two ``pre-place'' and ``placement'' frames, implies the desired movement of $T_\text{affordance}$ such that the object is placed. The movement from ``pre-place'' to ``placement'' is a simple top-down movement of some distance. The network jointly produce the $T_\text{affordance}$ and $T_\text{action}$ of both stages, i.e. the grasp pose produced by stage 1 should permit the execution of stage 2. This is connected to the idea of task-oriented grasp planning~\cite{tang2025foundationgrasp}.

\vspace{0.3em}
\noindent \textbf{Insert the cable into a clamp: } In this experiment, the robot needs to pick up a cable and vertically insert it into a clamp, as as shown in Fig.~\ref{fig:experiment} (b). We use 6 different ropes in this experiment. Both the rope and the clamp are randomly placed on the table. The rope, which is obviously deformable, is initially placed to different configurations.

This task has two stages. In stage 1, the robot would pick up the rope. In stage 2, the robot move the (grasped part of) rope to a suitable insert configuration and insert it into the clamp. As the grasp point predicted in stage 1 determines the ``controllable'' rope region in stage 2, these two stages must be predicted jointly by the network, else the robot cannot command the 6-DoF movements of the desired part of the rope. This further highlights the idea of task-oriented grasping mentioned above.

\vspace{0.3em}
\noindent \textbf{Hang the mug onto a rack: } In this experiment, the robot picks up a mug and hang it onto a rack, as shown in Fig.~\ref{fig:experiment} (a). We use 20 different mugs and 3 different racks in this experiment. This task requires center meter level accuracy of the entire manipulation pipeline, despite the shape variations of different objects.

This task has two stages. In stage 1, the robot picks up the mug. In stage 2, the robot move the mug to align its handle center with the stick of the rack. The $o_\text{manipulated}$ for this stage is the picked mug. The $T_\text{affordance}$ frame of stage 2 implies a configuration of the object that can be hung on a stick. The $T_\text{action\_seq}$ consists of three frames along the axis of the stick, with distance -3 [cm], 0 [cm], 2 [cm] relative to the tip of the stick. A visualization is shown in Fig~\ref{fig:action_output} (b), only one $T_\text{action}$ frame is visualized for clarity.


\begin{table}[t]
  \centering
 \includegraphics[width=0.8\linewidth]{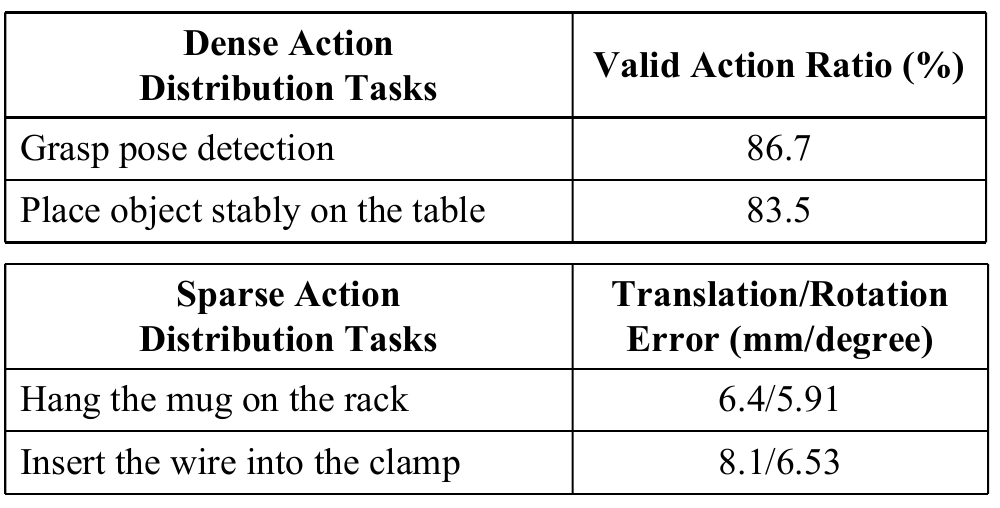}
 \caption{Analysis of the Generated Action Distribution}
 \label{table:successrate}
\end{table}

\subsection{Analysis of Generated Action Distribution}
\label{subsec:netoutput}

We perform an analysis of the generated action distribution for tasks mentioned above. A visualization of the generated action distribution is shown in Fig~\ref{fig:action_output}. For tasks with ``sparse'' action distribution where the ground-truth can be explicit computed, we present the translation and rotation error to the (closet) ground-truth action, as shown in Table.~\ref{table:successrate}. For tasks with ``dense'' action distribution that cannot be explicit computed, we report the success rate of the generated action, as shown in Table.~\ref{table:successrate}. For the grasp detection task, the success rate is computed using the pipeline methods in~\cite{urain2022se3}. For the stably placement task, a generate action is marked as success if: 1) the placement point is within 1[cm] from the table; 2) the z-axis of placement frame is aligned with the z-axis of the world frame within 15 degree; and 3) the placed object does not collide with any obstacles. It is emphasized that the overall success rate might be higher than valid action ratio, as the downstream trajectory generation can make selection from many action candidates generated by the network.

\begin{table}[t]
  \centering
 \includegraphics[width=0.8\linewidth]{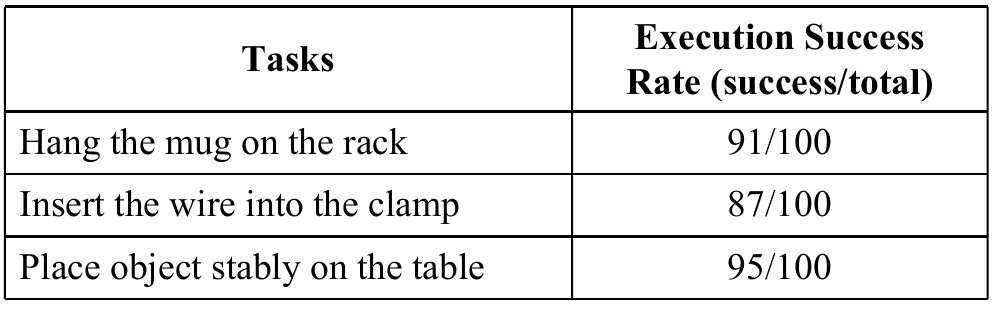}
 \caption{Statistic of the Overall Execution. It is emphasized that the overall success rate might be higher than valid action ratio, as the downstream can make selection from many action candidates generated by the network.}
 \label{table:overall_stat}
\end{table}

\subsection{Analysis of Overall Task Execution}
\label{subsec:netoutput}

The success rates of our method are summarized in Table~\ref{table:successrate}. For the cable-insertion task, we mark a trial as failure if the cable is not placed into the clamp. For the mug-on-rack task, we mark a trial as a failure if the mug is not hung onto the rack. For stable placement, we mark a trial as a failure if: 1) the trajectory collides with obstacles or the table; 2) the object falls down after placement. Our method achieves decent success rate despite the network needs to address multiple tasks and multiple objects \& configurations in each task. Fig.~\ref{fig:failure} illustrates several failure modes of our experiment.






\section{Conclusion}
\label{sec:conclusion}

In this paper, we propose a novel affordance-based formulation for end-to-end robotic manipulation. In particular, we model each manipulation task as task-specific oriented keypoints and their movements, both of them are predicted by a neural network. Our affordance-based formulation is used as the action representation of a end-to-end manipulation pipeline, which generalizes existing end-effector pose action representation. Many important manipulation tasks that were studied in isolation previously, e.g., robot grasping, can be expressed in a unified way with our formulation. The keypoint affordance enables natural generalization to objects with different shapes and sizes, while achieving sub-centimeter accuracy. Moreover, the proposed formulation can easily handle multi-stage tasks, multi-modality robot behaviors, and deformable objects. Extensive experiments demonstrate the effectiveness of our method.

{\small
\bibliographystyle{abbrv}
\bibliography{paper.bib}

\begin{thebibliography}{10}

\bibitem{achiam2023gpt}
J.~Achiam, S.~Adler, S.~Agarwal, L.~Ahmad, I.~Akkaya, F.~L. Aleman, D.~Almeida, J.~Altenschmidt, S.~Altman, S.~Anadkat, et~al.
\newblock Gpt-4 technical report.
\newblock {\em arXiv preprint arXiv:2303.08774}, 2023.

\bibitem{bai2023qwen}
J.~Bai, S.~Bai, Y.~Chu, Z.~Cui, K.~Dang, X.~Deng, Y.~Fan, W.~Ge, Y.~Han, F.~Huang, et~al.
\newblock Qwen technical report.
\newblock {\em arXiv preprint arXiv:2309.16609}, 2023.

\bibitem{black2024pi0}
K.~Black, N.~Brown, D.~Driess, A.~Esmail, M.~Equi, C.~Finn, N.~Fusai, L.~Groom, K.~Hausman, B.~Ichter, et~al.
\newblock Pi0: A vision-language-action flow model for general robot control.
\newblock {\em arXiv preprint arXiv:2410.24164}, 2024.

\bibitem{chi2023diffusion}
C.~Chi, Z.~Xu, S.~Feng, E.~Cousineau, Y.~Du, B.~Burchfiel, R.~Tedrake, and S.~Song.
\newblock Diffusion policy: Visuomotor policy learning via action diffusion.
\newblock {\em The International Journal of Robotics Research}, page 02783649241273668, 2023.

\bibitem{du2021vision}
G.~Du, K.~Wang, S.~Lian, and K.~Zhao.
\newblock Vision-based robotic grasping from object localization, object pose estimation to grasp estimation for parallel grippers: a review.
\newblock {\em Artificial Intelligence Review}, 54(3):1677--1734, 2021.

\bibitem{finn2016deep}
C.~Finn, X.~Y. Tan, Y.~Duan, T.~Darrell, S.~Levine, and P.~Abbeel.
\newblock Deep spatial autoencoders for visuomotor learning.
\newblock In {\em IEEE International Conference on Robotics and Automation (ICRA)}. IEEE, 2016.

\bibitem{fishman2023motion}
A.~Fishman, A.~Murali, C.~Eppner, B.~Peele, B.~Boots, and D.~Fox.
\newblock Motion policy networks.
\newblock In {\em conference on Robot Learning}, pages 967--977. PMLR, 2023.

\bibitem{gao2019kpam}
W.~Gao and R.~Tedrake.
\newblock kpam-sc: Generalizable manipulation planning using keypoint affordance and shape completion.
\newblock {\em arXiv preprint arXiv:1909.06980}, 2019.

\bibitem{gao2021kpam2}
W.~Gao and R.~Tedrake.
\newblock kpam 2.0: Feedback control for category-level robotic manipulation.
\newblock {\em IEEE Robotics and Automation Letters}, 6(2):2962--2969, 2021.

\bibitem{gao2024kilobot}
W.~Gao, J.~Wang, X.~Zhu, J.~Zhong, Y.~Shen, and Y.~Ding.
\newblock Kilobot: A programming language for deploying perception-guided industrial manipulators at scale.
\newblock {\em arXiv preprint arXiv:2409.03439}, 2024.

\bibitem{gualtieri2016gpd}
M.~Gualtieri, A.~Ten~Pas, K.~Saenko, and R.~Platt.
\newblock High precision grasp pose detection in dense clutter.
\newblock In {\em 2016 IEEE/RSJ International Conference on Intelligent Robots and Systems (IROS)}, pages 598--605. IEEE, 2016.

\bibitem{huang2024rekep}
W.~Huang, C.~Wang, Y.~Li, R.~Zhang, and L.~Fei-Fei.
\newblock Rekep: Spatio-temporal reasoning of relational keypoint constraints for robotic manipulation.
\newblock {\em arXiv preprint arXiv:2409.01652}, 2024.

\bibitem{ke20243d}
T.-W. Ke, N.~Gkanatsios, and K.~Fragkiadaki.
\newblock 3d diffuser actor: Policy diffusion with 3d scene representations.
\newblock {\em arXiv preprint arXiv:2402.10885}, 2024.

\bibitem{kim2024openvla}
M.~J. Kim, K.~Pertsch, S.~Karamcheti, T.~Xiao, A.~Balakrishna, S.~Nair, R.~Rafailov, E.~Foster, G.~Lam, P.~Sanketi, et~al.
\newblock Openvla: An open-source vision-language-action model.
\newblock {\em arXiv preprint arXiv:2406.09246}, 2024.

\bibitem{kokic2017affordance}
M.~Kokic, J.~A. Stork, J.~A. Haustein, and D.~Kragic.
\newblock Affordance detection for task-specific grasping using deep learning.
\newblock In {\em 2017 IEEE-RAS 17th International Conference on Humanoid Robotics (Humanoids)}, pages 91--98. IEEE, 2017.

\bibitem{levine2016end}
S.~Levine, C.~Finn, T.~Darrell, and P.~Abbeel.
\newblock End-to-end training of deep visuomotor policies.
\newblock {\em The Journal of Machine Learning Research}, 17(1):1334--1373, 2016.

\bibitem{levine2016contact}
S.~Levine, N.~Wagener, and P.~Abbeel.
\newblock Learning contact-rich manipulation skills with guided policy search.
\newblock {\em arXiv preprint}, 2016.

\bibitem{li2024laso}
Y.~Li, N.~Zhao, J.~Xiao, C.~Feng, X.~Wang, and T.-s. Chua.
\newblock Laso: Language-guided affordance segmentation on 3d object.
\newblock In {\em Proceedings of the IEEE/CVF Conference on Computer Vision and Pattern Recognition}, pages 14251--14260, 2024.

\bibitem{liang2022code}
J.~Liang, W.~Huang, F.~Xia, P.~Xu, K.~Hausman, B.~Ichter, P.~Florence, and A.~Zeng.
\newblock Code as policies: Language model programs for embodied control.
\newblock {\em arXiv preprint arXiv:2209.07753}, 2022.

\bibitem{liu2024improved}
H.~Liu, C.~Li, Y.~Li, and Y.~J. Lee.
\newblock Improved baselines with visual instruction tuning.
\newblock In {\em Proceedings of the IEEE/CVF conference on computer vision and pattern recognition}, pages 26296--26306, 2024.

\bibitem{liu2023visual}
H.~Liu, C.~Li, Q.~Wu, and Y.~J. Lee.
\newblock Visual instruction tuning.
\newblock {\em Advances in neural information processing systems}, 36:34892--34916, 2023.

\bibitem{liu2024rdt}
S.~Liu, L.~Wu, B.~Li, H.~Tan, H.~Chen, Z.~Wang, K.~Xu, H.~Su, and J.~Zhu.
\newblock Rdt-1b: a diffusion foundation model for bimanual manipulation.
\newblock {\em arXiv preprint arXiv:2410.07864}, 2024.

\bibitem{ma2024groma}
C.~Ma, Y.~Jiang, J.~Wu, Z.~Yuan, and X.~Qi.
\newblock Groma: Localized visual tokenization for grounding multimodal large language models.
\newblock In {\em European Conference on Computer Vision}, pages 417--435. Springer, 2024.

\bibitem{mahler2019learning}
J.~Mahler, M.~Matl, V.~Satish, M.~Danielczuk, B.~DeRose, S.~McKinley, and K.~Goldberg.
\newblock Learning ambidextrous robot grasping policies.
\newblock {\em Science Robotics}, 4(26):eaau4984, 2019.

\bibitem{morrison2018closing}
D.~Morrison, P.~Corke, and J.~Leitner.
\newblock Closing the loop for robotic grasping: A real-time, generative grasp synthesis approach.
\newblock {\em arXiv preprint arXiv:1804.05172}, 2018.

\bibitem{myers2015affordance}
A.~Myers, C.~L. Teo, C.~Ferm{\"u}ller, and Y.~Aloimonos.
\newblock Affordance detection of tool parts from geometric features.
\newblock In {\em 2015 IEEE international conference on robotics and automation (ICRA)}, pages 1374--1381. IEEE, 2015.

\bibitem{noreen2016optimal}
I.~Noreen, A.~Khan, and Z.~Habib.
\newblock Optimal path planning using rrt* based approaches: a survey and future directions.
\newblock {\em International Journal of Advanced Computer Science and Applications}, 7(11), 2016.

\bibitem{osiurak2017affordance}
F.~Osiurak, Y.~Rossetti, and A.~Badets.
\newblock What is an affordance? 40 years later.
\newblock {\em Neuroscience \& Biobehavioral Reviews}, 77:403--417, 2017.

\bibitem{pan2025omnimanip}
M.~Pan, J.~Zhang, T.~Wu, Y.~Zhao, W.~Gao, and H.~Dong.
\newblock Omnimanip: Towards general robotic manipulation via object-centric interaction primitives as spatial constraints.
\newblock In {\em Proceedings of the Computer Vision and Pattern Recognition Conference}, pages 17359--17369, 2025.

\bibitem{qin2019keto}
Z.~Qin, K.~Fang, Y.~Zhu, L.~Fei-Fei, and S.~Savarese.
\newblock Keto: Learning keypoint representations for tool manipulation.
\newblock {\em arXiv preprint arXiv:1910.11977}, 2019.

\bibitem{sahin2018category}
C.~Sahin and T.-K. Kim.
\newblock Category-level 6d object pose recovery in depth images.
\newblock In {\em European Conference on Computer Vision}, pages 665--681. Springer, 2018.

\bibitem{schulman2015trust}
J.~Schulman, S.~Levine, P.~Abbeel, M.~Jordan, and P.~Moritz.
\newblock Trust region policy optimization.
\newblock In {\em International conference on machine learning}, pages 1889--1897, 2015.

\bibitem{schulman2017proximal}
J.~Schulman, F.~Wolski, P.~Dhariwal, A.~Radford, and O.~Klimov.
\newblock Proximal policy optimization algorithms.
\newblock {\em arXiv:1707.06347}, 2017.

\bibitem{tang2025foundationgrasp}
C.~Tang, D.~Huang, W.~Dong, R.~Xu, and H.~Zhang.
\newblock Foundationgrasp: Generalizable task-oriented grasping with foundation models.
\newblock {\em IEEE Transactions on Automation Science and Engineering}, 2025.

\bibitem{tang2025uad}
Y.~Tang, W.~Huang, Y.~Wang, C.~Li, R.~Yuan, R.~Zhang, J.~Wu, and L.~Fei-Fei.
\newblock Uad: Unsupervised affordance distillation for generalization in robotic manipulation.
\newblock {\em arXiv preprint arXiv:2506.09284}, 2025.

\bibitem{touvron2023llama}
H.~Touvron, T.~Lavril, G.~Izacard, X.~Martinet, M.-A. Lachaux, T.~Lacroix, B.~Rozi{\`e}re, N.~Goyal, E.~Hambro, F.~Azhar, et~al.
\newblock Llama: Open and efficient foundation language models.
\newblock {\em arXiv preprint arXiv:2302.13971}, 2023.

\bibitem{tremblay2018deep}
J.~Tremblay, T.~To, B.~Sundaralingam, Y.~Xiang, D.~Fox, and S.~Birchfield.
\newblock Deep object pose estimation for semantic robotic grasping of household objects.
\newblock {\em Conference on Robot Learning (CoRL)}, 2018.

\bibitem{urain2022se3}
J.~Urain, N.~Funk, J.~Peters, and G.~Chalvatzaki.
\newblock Se (3)-diffusionfields: Learning smooth cost functions for joint grasp and motion optimization through diffusion.
\newblock {\em arXiv preprint arXiv:2209.03855}, 2022.

\bibitem{van2016stable}
H.~Van~Hoof, N.~Chen, M.~Karl, P.~van~der Smagt, and J.~Peters.
\newblock Stable reinforcement learning with autoencoders for tactile and visual data.
\newblock In {\em International Conference on Intelligent Robots and Systems}, 2016.

\bibitem{wang2024qwen2vl}
P.~Wang, S.~Bai, S.~Tan, S.~Wang, Z.~Fan, J.~Bai, K.~Chen, X.~Liu, J.~Wang, W.~Ge, et~al.
\newblock Qwen2-vl: Enhancing vision-language model's perception of the world at any resolution.
\newblock {\em arXiv preprint arXiv:2409.12191}, 2024.

\bibitem{wang2025skil}
S.~Wang, J.~You, Y.~Hu, J.~Li, and Y.~Gao.
\newblock Skil: Semantic keypoint imitation learning for generalizable data-efficient manipulation.
\newblock {\em arXiv preprint arXiv:2501.14400}, 2025.

\bibitem{wang2023robogen}
Y.~Wang, Z.~Xian, F.~Chen, T.-H. Wang, Y.~Wang, K.~Fragkiadaki, Z.~Erickson, D.~Held, and C.~Gan.
\newblock Robogen: Towards unleashing infinite data for automated robot learning via generative simulation.
\newblock {\em arXiv preprint arXiv:2311.01455}, 2023.

\bibitem{wong2021manipulation}
C.-C. Wong, L.-Y. Yeh, C.-C. Liu, C.-Y. Tsai, and H.~Aoyama.
\newblock Manipulation planning for object re-orientation based on semantic segmentation keypoint detection.
\newblock {\em Sensors}, 21(7):2280, 2021.

\bibitem{zawalski2024robotic}
M.~Zawalski, W.~Chen, K.~Pertsch, O.~Mees, C.~Finn, and S.~Levine.
\newblock Robotic control via embodied chain-of-thought reasoning.
\newblock {\em arXiv preprint arXiv:2407.08693}, 2024.

\bibitem{zeng2018affordance}
A.~Zeng, S.~Song, K.-T. Yu, E.~Donlon, F.~R. Hogan, M.~Bauza, D.~Ma, O.~Taylor, M.~Liu, E.~Romo, et~al.
\newblock Robotic pick-and-place of novel objects in clutter with multi-affordance grasping and cross-domain image matching.
\newblock In {\em 2018 IEEE International Conference on Robotics and Automation (ICRA)}, pages 1--8. IEEE, 2018.

\bibitem{zeng2017multi}
A.~Zeng, K.-T. Yu, S.~Song, D.~Suo, E.~Walker, A.~Rodriguez, and J.~Xiao.
\newblock Multi-view self-supervised deep learning for 6d pose estimation in the amazon picking challenge.
\newblock In {\em 2017 IEEE International Conference on Robotics and Automation (ICRA)}, pages 1386--1383. IEEE, 2017.

\bibitem{zhao2023act}
T.~Z. Zhao, V.~Kumar, S.~Levine, and C.~Finn.
\newblock Learning fine-grained bimanual manipulation with low-cost hardware.
\newblock {\em arXiv preprint arXiv:2304.13705}, 2023.

\bibitem{zhu2018reinforcement}
Y.~Zhu, Z.~Wang, J.~Merel, A.~Rusu, T.~Erez, S.~Cabi, S.~Tunyasuvunakool, J.~Kram{\'a}r, R.~Hadsell, N.~de~Freitas, et~al.
\newblock Reinforcement and imitation learning for diverse visuomotor skills.
\newblock {\em arXiv preprint}, 2018.

\end{thebibliography}
}

\end{document}